\newcommand{\CLASSINPUTbaselinestretch}{1.0} % baselinestretch
\newcommand{\CLASSINPUTinnersidemargin}{0.5in} % inner side margin
\newcommand{\CLASSINPUToutersidemargin}{0.5in} % outer side margin
\newcommand{\CLASSINPUTtoptextmargin}{0.5in}   % top text margin
\newcommand{\CLASSINPUTbottomtextmargin}{0.5in}% bottom text margin

\documentclass[10pt, technote,compsoc]{IEEEtran}
\usepackage[cmex10]{amsmath}
\usepackage{fixltx2e}
\usepackage{datetime}
\usepackage{graphicx}
\usepackage{color}
\usepackage{amssymb}
\usepackage{amssymb}

\usepackage{pstricks,epsfig}
\usepackage{rotating}
\usepackage{subfig}
\usepackage{tikz}
\usepackage{enumerate}
\usepackage{tabularx}
\usepackage{pgf}
\usepackage{url}
\usepackage{color,soul}

\ifCLASSOPTIONcompsoc
  \usepackage[nocompress]{cite}
\else
  \usepackage{cite}
\fi

\ifCLASSINFOpdf

\else

\fi

\begin{document}

\title{Efficient Deep Feature Learning and Extraction via StochasticNets}

\author{Mohammad~Javad~Shafiee,
        Parthipan~Siva
        Paul~Fieguth
        and~Alexander~Wong% <-this % stops a space
\IEEEcompsocitemizethanks{\IEEEcompsocthanksitem The authors are with the Department of Systems Design Engineering, University of Waterloo, Waterloo, Ontario, Canada.\protect\\
E-mail: \{mjshafiee, a28wong, pfieguth\}@ uwaterloo.ca}% <-this % stops a space
\thanks{Manuscript received ..., 2015; revised ... .}}

%\markboth{IEEE TRANSACTIONS ON PATTERN ANALYSIS AND MACHINE INTELLIGENCE,~Vol.~?, No.~?, June~?}%
%{Shell \MakeLowercase{\textit{et al.}}: Bare Advanced Demo of IEEEtran.cls for Journals}

\IEEEtitleabstractindextext{%
\begin{abstract}
Deep neural networks are a powerful tool for feature learning and extraction given their ability to model high-level abstractions in highly complex data. One area worth exploring in feature learning and extraction using deep neural networks is efficient neural connectivity formation for faster feature learning and extraction. Motivated by findings of stochastic synaptic connectivity formation in the brain as well as the brain's uncanny ability to efficiently represent information, we propose the efficient learning and extraction of features via StochasticNets, where sparsely-connected deep neural networks can be formed via stochastic connectivity between neurons. To evaluate the feasibility of such a deep neural network architecture for feature learning and extraction, we train deep convolutional StochasticNets to learn abstract features using the CIFAR-10 dataset, and extract the learned features from images to perform classification on the SVHN and STL-10 datasets. Experimental results show that features learned using deep convolutional StochasticNets, with fewer neural connections than conventional deep convolutional neural networks, can allow for better or comparable classification accuracy than conventional deep neural networks: relative test error decrease of $\sim$4.5\% for classification on the \mbox{STL-10} dataset and $\sim$1\% for classification on the SVHN dataset. Furthermore, it was shown that the deep features extracted using deep convolutional StochasticNets can provide comparable classification accuracy even when only 10\% of the training data is used for feature learning.  Finally, it was also shown that significant gains in feature extraction speed can be achieved in embedded applications using StochasticNets.  As such, StochasticNets allow for faster feature learning and extraction performance while facilitate for better or comparable accuracy performances.
\end{abstract}

}

\maketitle

\IEEEdisplaynontitleabstractindextext
\IEEEpeerreviewmaketitle

%\vspace{-0.35 cm}
 \section{Introduction}

Deep neural networks are a powerful tool for feature learning and extraction given their ability to represent and model high-level abstractions in highly complex data.  Deep neural networks have shown considerable capabilities in producing features that enable state-of-the-art performance for handling complex tasks such as speech recognition~\cite{Hannun,Dahl}, object recognition~\cite{Krizhevsky,He,LeCun,Simonyan}, and natural language processing~\cite{Collobert,Bengio}.  Recent advances in improving the performance of deep neural networks for feature learning and extraction have focused on areas such as network regularization~\cite{Zeller,Wan}, activation functions~\cite{Glorot1,Glorot2,He2}, and deeper architectures~\cite{Simonyan,Szegedy,Zhang}, where the goal is to learn more representative features with respect to increasing task accuracy.

Despite the power capabilities of deep neural networks for feature learning and extraction, they are very rarely employed on embedded devices such as video surveillance cameras, smartphones, and wearable devices.  This difficult migration of deep neural networks into embedded applications for feature extraction stems largely from the fact that, unlike the highly powerful distributed computing systems and GPUs that are often leveraged for deep learning networks,  the low-power CPUs commonly used in embedded systems simply do not have the computational power to make deep neural networks a feasible solution for feature extraction.

Much of the focus on migrating deep neural networks for feature learning and extraction in embedded systems have been to design custom embedded processing units dedicated to accelerating deep neural networks~\cite{hardwareCNN_2010_Farbet,MobileCNN_2014_Jin,CNNProcessor_2014_Gokhale} .  However, such an approach greatly limits the flexibility of the type of deep neural network architectures that can be used.  Furthermore, such an approach requires adding additional hardware, which adds to the cost and complexity of the embedded system.  On the other hand, improving the efficiency of deep neural networks for feature learning and extraction is much less explored, with considerably fewer strategies proposed so far~\cite{HashNet_2015Wenlin}.  In particular, very little exploration has been conducted on efficient neural connectivity formation for efficient feature learning and extraction, which can hold considerable promise it achieving highly efficient deep neural network architectures that can be used in embedded applications.

One way to address this challenge is to draw inspiration from the brain which has an uncanny ability to efficiently represent information.  In particular, we are inspired by the way brain develops synaptic connectivity between neurons.  Recently, in a pivotal paper by~\cite{Hill}, data of living brain tissue from Wistar rats was collected and used to construct a partial map of a rat brain.  Based on this map, \mbox{Hill et al.} came to a very surprising conclusion. The synaptic formation, of specific functional connectivity in neocortical neural microcircuits, was found to be stochastic in nature. This is in sharp contrast to the way deep neural networks are formed, where connectivity is largely deterministic and pre-defined.
\begin{figure}

\begin{center}
\includegraphics[scale = 0.45]{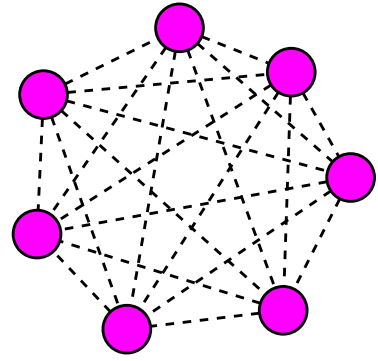}
\caption{An illustrative example of a random graph. All possible edge connectivity between the nodes in the graph may occur independently with a probability of $p_{ij}$.  }
\label{fig:RG}
\end{center}
%%\vspace{-1 cm}
\end{figure}

Motivated by findings of random neural connectivity formation and the efficient information representation capabilities of the brain, we proposed the learning of efficient feature representations via StochasticNets~\cite{stochasticnet}, where the key idea is to leverage random graph theory~\cite{Gilbert,Erdos} to form sparsely-connected deep neural networks via stochastic connectivity between neurons. The connection sparsity, in particular for deep convolutional networks, allows for more efficient feature learning and extraction due to the sparse nature of receptive fields which requires less computation and memory access. We will show that these sparsely-connected deep neural networks, while computationally efficient, can still maintain the same accuracies as the traditional deep neural networks.  Furthermore, the StochasticNet architecture for feature learning and extraction presented in this work can also benefit from all of the same approaches used for traditional deep neural networks such as data augmentation and stochastic pooling to further improve performance.

The paper is organized as follows. First, a review of random graph theory is presented in Section~\ref{sec:randomgraph}.  The theory and design considerations behind forming StochasticNet as a random graph realizations are discussed in Section 3.  Experimental results where we train deep convolutional StochasticNets to learn abstract features using the CIFAR-10 dataset~\cite{CIFAR10}, and extract the learned features from images to perform classification on the SVHN~\cite{SVHN} and STL-10~\cite{STL10} datasets is presented in Section 5.  Finally, conclusions are drawn in Section 6.

 %\vspace{- 0.25 cm}
\section{Review of Random Graph Theory}
\label{sec:randomgraph}
The underlying idea of deep feature learning via StochasticNets is to leverage random graph theory~\cite{Gilbert,Erdos} to form the neural connectivity of deep neural networks in a stochastic manner such that the resulting neural networks are sparsely connected yet maintaining feature representation capabilities.  As such, it is important to first provide a general overview of random graph theory for context.  In random graph theory, a random graph can be defined as the probability distribution over graphs~\cite{Bollobas}.  A number of different random graph models have been proposed in literature.

A commonly studied random graph model is that proposed by~\cite{Gilbert}, in which a random graph can be expressed by $\mathcal{G}(n,p)$, where all possible edge connectivity are said to occur independently with a probability of $p$, where $0 < p < 1$.  This random graph model was generalized by~\cite{Kovalenko}, in which a random graph can be expressed by $\mathcal{G}(\mathcal{V},p_{ij})$, where $\mathcal{V}$ is a set of vertices and the edge connectivity between two vertices $\{i,j\}$ in the graph is said to occur with a probability of $p_{ij}$, where $p_{ij} \in[0, 1]$.

\begin{figure*}
\begin{center}
\begin{tabular}{cccc}
\includegraphics[scale = 0.3]{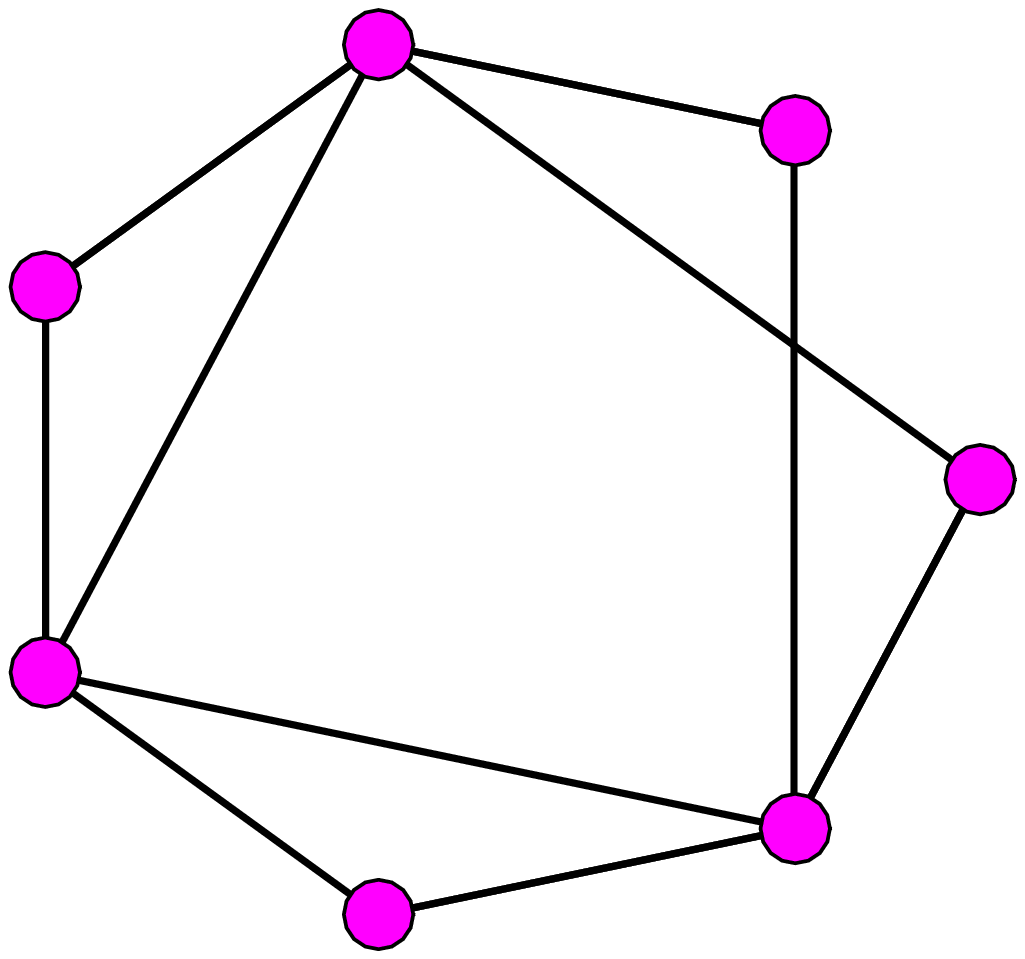}&
\includegraphics[scale = 0.3]{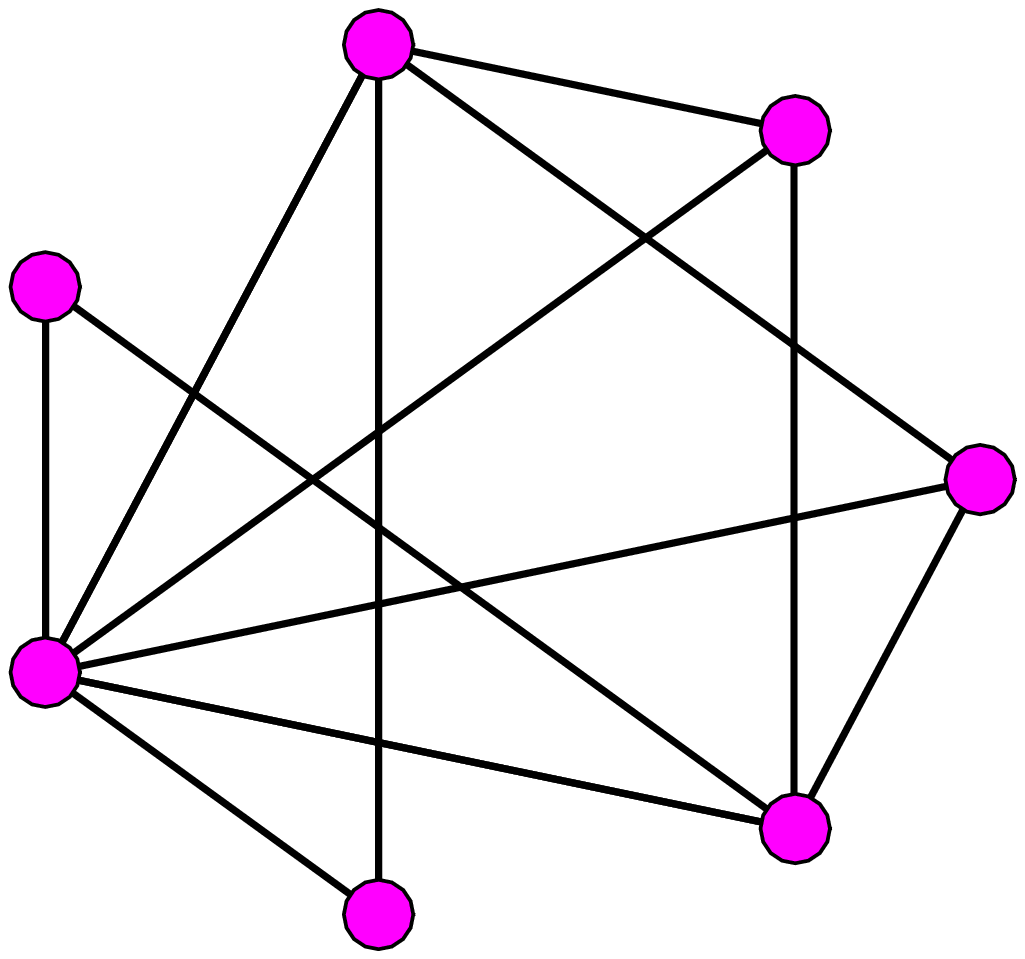}&
\includegraphics[scale = 0.3]{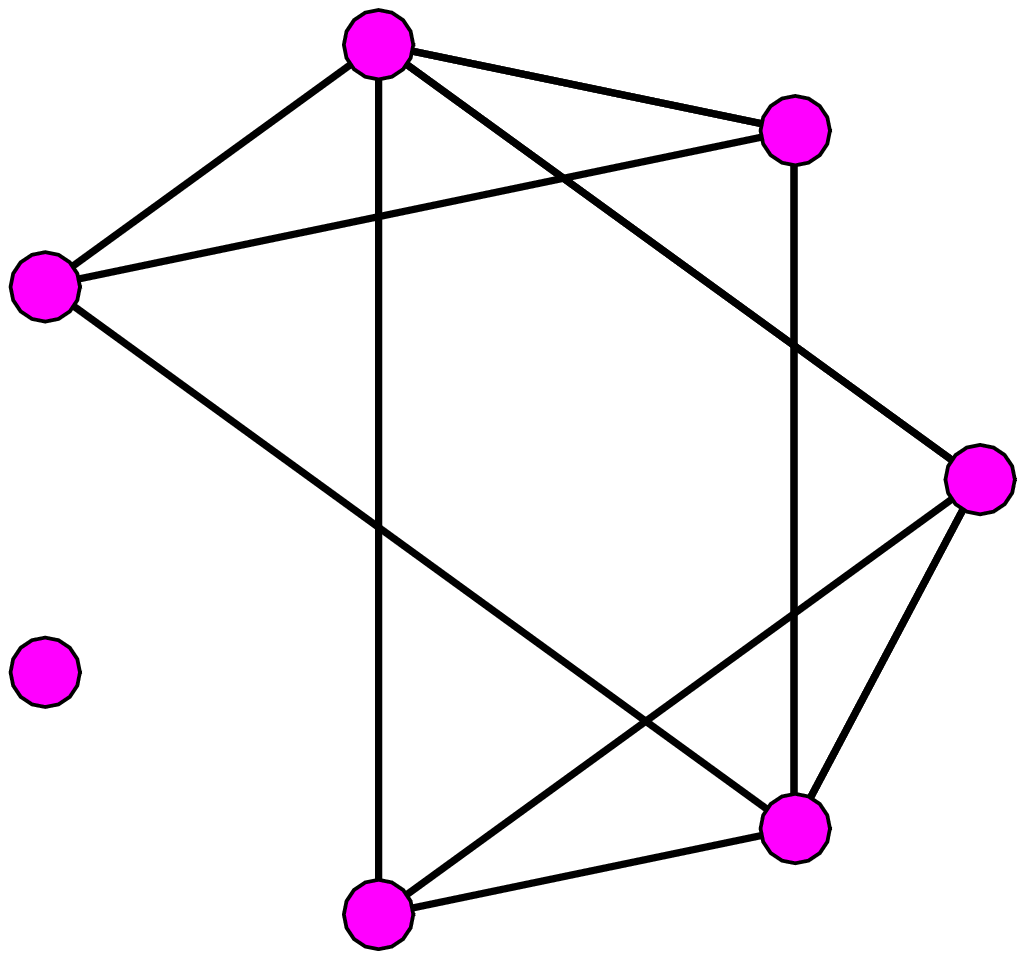}&
\includegraphics[scale = 0.3]{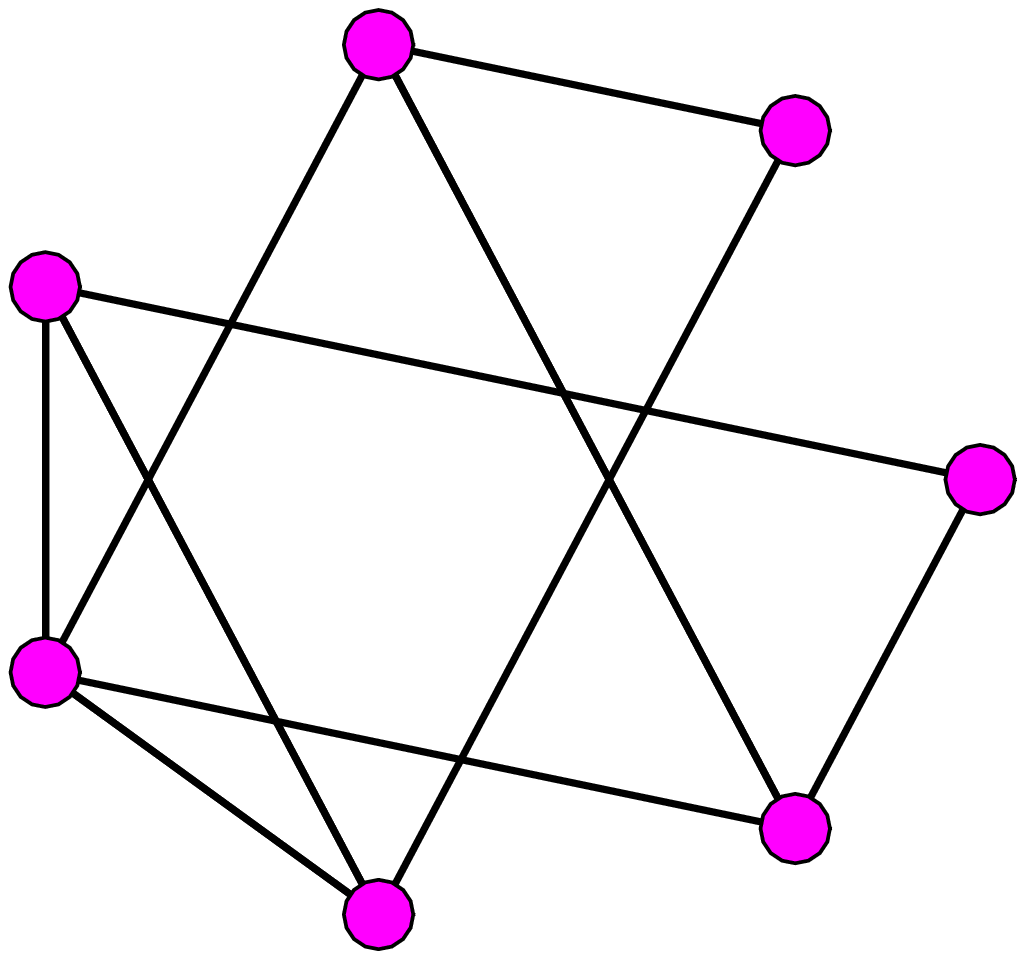}
\end{tabular}

\caption{ Realizations of a random graph with 50 nodes. The probability for edge connectivity between all nodes in the graph was set to $p_{i,j}=0.5$ for all nodes $i$ and $j$. Each diagram demonstrates a different realization of the random graph. }
\label{fig:RGrealization}
\end{center}
%\vspace{-0.5 cm}
\end{figure*}

Therefore, based on this generalized random graph model, realizations of random graphs can be obtained by starting with a set of $n$ vertices $\mathcal{V} = \{v_q|1 \leq q \leq n\}$ and randomly adding a set of edges between the vertices based on the set of possible edges \mbox{$\mathcal{E} = \{e_{ij}|1 \leq i \leq n, 1 \leq j \leq n\ , i  \neq j \}$} independently with a probability of $p_{ij}$.   A number of realizations of a random graph are provided in Figure~\ref{fig:RGrealization} for illustrative purposes. It is worth noting that because of the underlying probability distribution, the generated realizations of the random graph often exhibit differing edge connectivity.

Given that deep neural networks can be fundamentally expressed and represented as graphs $\mathcal{G}$, where the neurons are vertices $\mathcal{V}$ and the neural connections are edges $\mathcal{E}$, one intriguing idea for introducing stochastic connectivity for the formation of deep neural networks is to treat the formation of deep neural networks as particular realizations of random graphs, which we will describe in greater detail in the next section.

 %\vspace{- 0.25 cm}
\section{StochasticNets: Deep Neural Networks as Random Graph Realizations}

\begin{figure}[!htp]
\begin{center}
\begin{tabular}{cc}
\includegraphics[width = 4.25 cm]{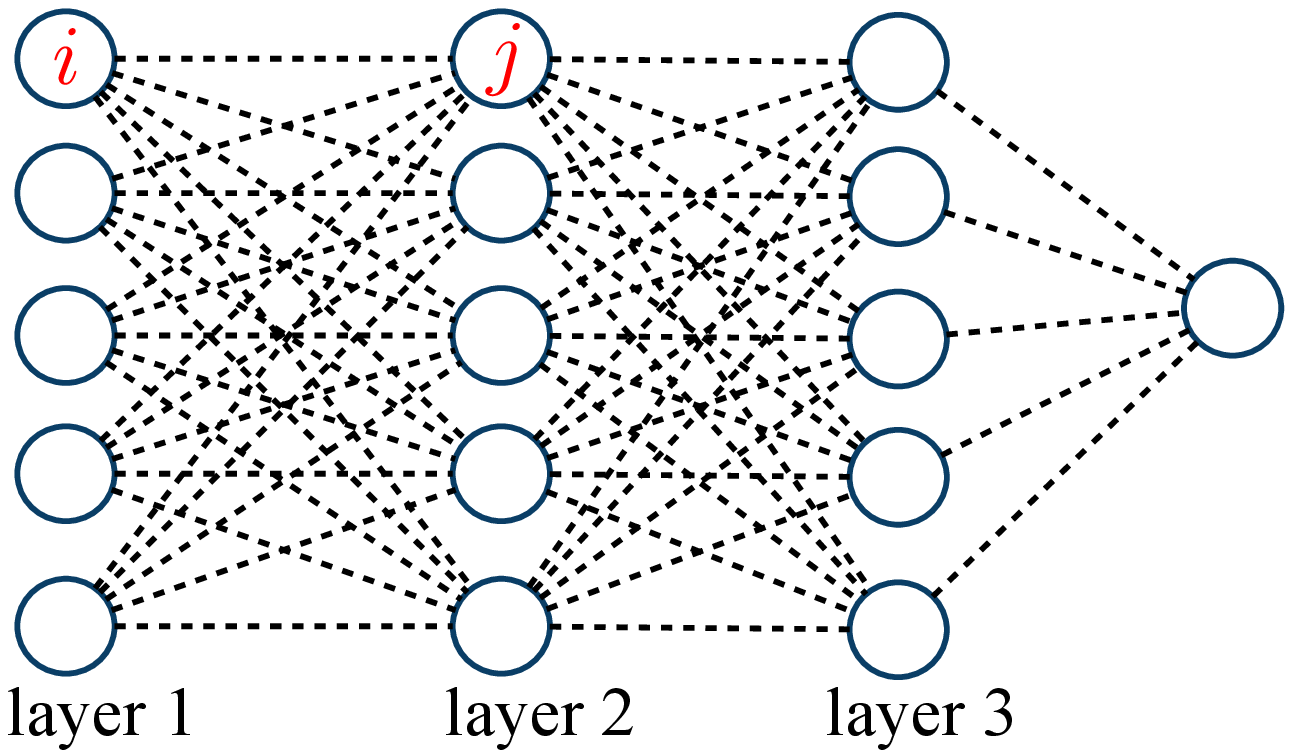}&
\includegraphics[width = 4.25 cm]{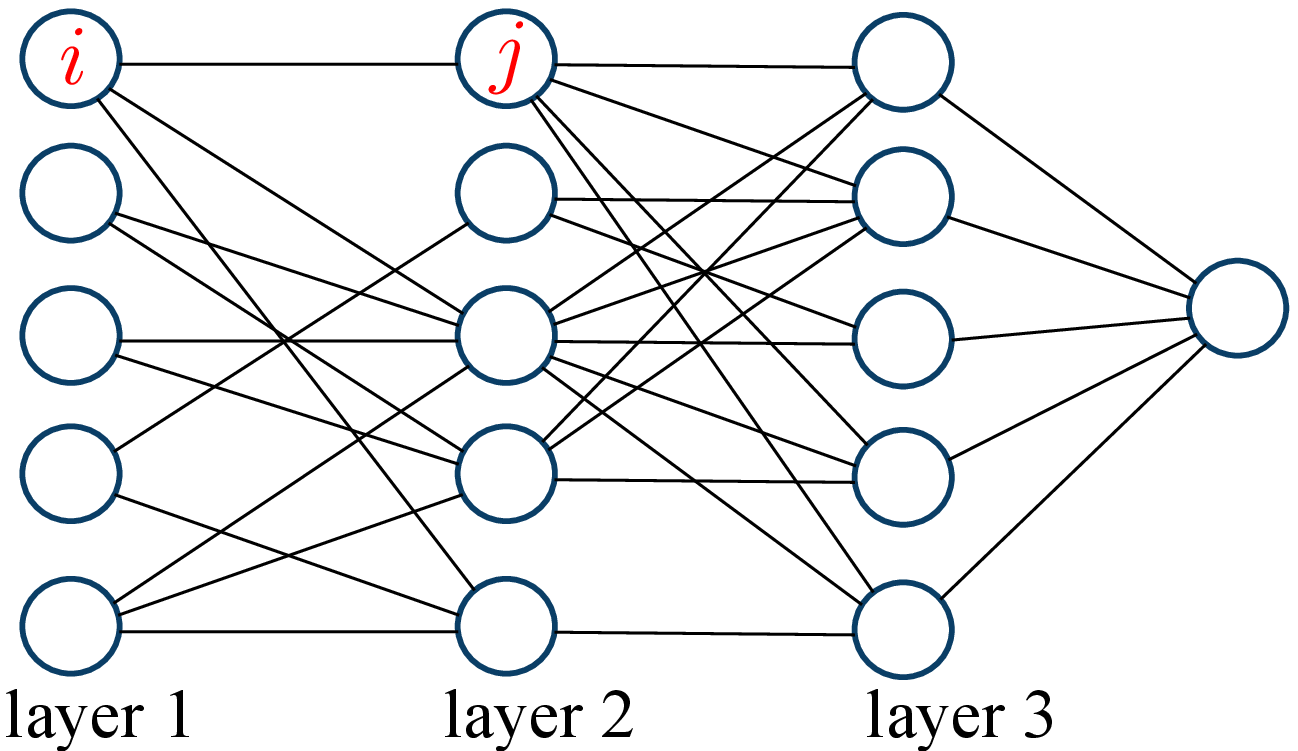}\\
(a) RG Representation & (b) FFNN Graph Realization
\end{tabular}
\caption{ Example random graph representing a section of a deep feed-forward neural network (a) and its realization (b). Every neuron $i$  may be connected to neuron $j$ with probability $p({i\rightarrow j})$ based on random graph theory.  To enforce the properties of a general deep feed-forward neural network, there is no neural  connectivity between nodes that are not in the adjacent layers. As shown in (b), the neural  connectivity of nodes in the realization are varied since they are drawn based on a probability distribution. }
\label{fig:StochasticNetRG}
\end{center}
%\vspace{-0.5 cm}
\end{figure}

Let us represent a deep neural network as a random graph \mbox{$\mathcal{G}\Big(\mathcal{V},p({i\rightarrow j})\Big)$}, where $\mathcal{V}$ is the set of neurons \mbox{$\mathcal{V} = \{v_{i}|1 \geq i \geq n\}$}, with $v_{i}$ denoting the $i^{\rm th}$ neuron and $n$ denoting the total number of neurons, in the deep neural network and \mbox{$p({i\rightarrow j})$} is the probability that a neural connection occurs between neuron $v_{i}$ and $v_{j}$. It is worth noting that since neural networks are directed graphs, to this end the probability $p(\cdot)$ is represented in a directed way from source node to the destination node.  As such, one can then form a deep neural network as a realization of the random graph \mbox{$\mathcal{G}\Big(\mathcal{V},p({i\rightarrow j})\Big)$} by starting with a set of neurons $\mathcal{V}$, and randomly adding neural connections between the set of neurons independently with a probability of $p({i\rightarrow j})$ as defined above.

While one can form practically any type of deep neural network as a random graph realizations, an important design consideration for forming deep neural networks as random graph realizations is that different types of deep neural networks have fundamental properties in their network architecture that must be taken into account and preserved in the random graph realization.  Therefore, to ensure that fundamental properties of the network architecture of a certain type of deep neural network is preserved, the probability $p({i\rightarrow j})$ must be designed in such a way that these properties are enforced appropriately in the resultant random graph realization.  Let us consider a general deep feed-forward neural network.  First, in a deep feed-forward neural network, there can be no neural connections between non-adjacent layers.  Second, in a deep feed-forward neural network, there can be no neural connections between neurons on the same layer.  Therefore, to enforce these two properties, $p({i\rightarrow j}) = 0 $ when $l(i) \neq l(j) + 1$ where $l(i)$ encodes the layer number associated to the node $i$.  An example random graph based on this random graph model for representing general deep feed-forward neural networks is shown in Figure~\ref{fig:StochasticNetRG}(a), with an example feed-forward neural network graph realization is shown in Figure~\ref{fig:StochasticNetRG}(b).  Note that the neural connectivity for each neuron may be different due to the stochastic nature of neural connection formation.
 %\vspace{- 0.25 cm}
\section{Feature Learning via Deep Convolutional StochasticNets}
As one of the most commonly used types of deep neural networks for feature learning is deep convolutional neural networks~\cite{MNIST}, let us investigate the efficacy of efficient feature learning can be achieved via deep convolutional StochasticNets.
Deep convolutional neural networks can provide a general  abstraction of the input data by applying the sequential convolutional layers to the input data (e.g. input image). The goal of convolutional layers in a deep neural network is to extract  discriminative features to feed into the classifier such that the fully connected layers play the role of classification in deep neural networks. Therefore, the combination of receptive fields in convolutional layers can be considered as the feature extractor in these models. The receptive fields' parameters must be trained to find optimal parameters leading to most discriminative features.

However learning those parameters  is not possible every time due to the computational complexity or lake of enough training data such that  general receptive fields (i.e., convolutional layers) without learning is desirable. On the other hand, the computational complexity of extracting features is another concern which should be addressed. Essentially, extracting features in a deep convolutional neural network is a sequence of convolutional processes which can be represented as multiplications and summations and the number of operations is dependent on the number of parameters of receptive fields. Motivated by those reasons, sparsifying the  receptive fields while maintaining the generality of them is highly desirable.

To this end, we want to sparsify the receptive field motivated by the StochasticNet framework to provide efficient deep feature learning.
First, in additional to the design considerations for $p({i\rightarrow j})$ presented in the previous section to enforce the properties of deep feed-forward neural networks, additional considerations must be taken to preserve the properties of deep convolutional neural networks, which is a type of deep feed-forward neural network.

\begin{figure}[!htp]
\begin{center}
\includegraphics[scale = 0.5]{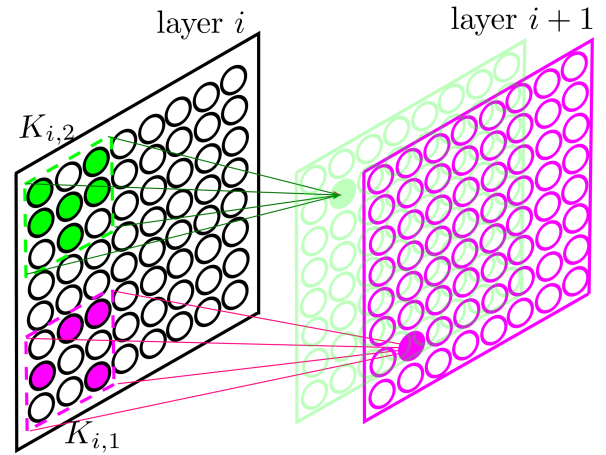}
\caption{ Forming a deep convolutional StochasticNet.  The neural connectivity for each randomly realized receptive field (in this case, $\{K_{i,1},K_{i,2}\}$) are determined based on a probability distribution, and as such the configuration of each randomly realized  receptive field may differ.  It can be seen that the shape and neural connectivity for receptive field $K_{i,1}$ is completely different from  receptive field $K_{i,2}$.  The response of each randomly realized receptive field leads to an output in new channel $C$. Only one layer of the formed deep convolutional StochasticNet is shown for illustrative purposes.   }
\label{fig:StochasticCNN}
\end{center}
\end{figure}

Specifically, the neural connectivity for each randomly realized receptive field $K$ in the deep convolutional StochasticNet is based on a probability distribution, with the neural connectivity configuration thus being shared amongst different small neural collections for a given randomly realized  receptive field.  An example of a realized deep convolutional StochasticNet is shown in Figure~\ref{fig:StochasticCNN}.  As seen, the neural connectivity for randomly realized receptive field $K_{i,1}$ is completely different from randomly realized  receptive field $K_{i,2}$.  The response of each randomly realized receptive field leads to an output in new channel in layer $i+1$.

A realized deep convolutional StochasticNet can then be trained to learn efficient feature representations via supervised learning using labeled data.  One can then use the trained StochasticNet for extracting a set of abstract features from input data.

 \subsection{Relationship to Other Methods}
While a number of stochastic strategies for improving deep neural network performance have been previously introduced~\cite{Wan}, \cite{HashNet_2015Wenlin} and \cite{Dropout_2014_Nitish}, it is very important to note that the proposed StochasticNets is fundamentally different from these existing stochastic strategies in that StochasticNets’ main significant contributions deals primarily with the formation of neural connectivity of individual neurons  to construct efficient deep neural networks that are inherently sparse prior to training, while previous stochastic strategies deal with either the grouping of existing neural connections to explicitly enforce sparsity~\cite{HashNet_2015Wenlin}, or removal/introduction of neural connectivity for regularization during training. More specifically, StochasticNets is a realization of a random graph formed prior to training and as such the connectivity in the network are inherently sparse, and are permanent and do not change during training. This is very different from Dropout~\cite{Dropout_2014_Nitish} and DropConnect~\cite{Wan} where the activations and connections are temporarily  removed during training and put back during test for regularization purposes only, and as such the resulting neural connectivity of the network remains dense. There is no notion of ’dropping’ in StochasticNets as only a subset of possible neural connections are formed in the first place prior to training, and the resulting network connectivity of the network is sparse.

StochasticNets are also very different from HashNets~\cite{HashNet_2015Wenlin}, where connection weights are randomly grouped into hash buckets,
with each bucket sharing the same weights, to explicitly sparsifying into the network, since there is no notion of grouping/merging in StochasticNets; the formed StochasticNets are naturally sparse due to the formation process. In fact, stochastic strategies such as HashNets, Dropout, and DropConnect can be used in conjunction with StochasticNets.
 \begin{figure*}
\vspace{- 0.75cm}
\begin{center}
\begin{tabular}{c c|c|c|c}
~& Network Model & Network Formation & Training & Testing  \\\hline \hline

  \raisebox{13 pt}[0 pt][0 pt]{\rotatebox{90}{\footnotesize Dropout}}&
\includegraphics[width = 4 cm]{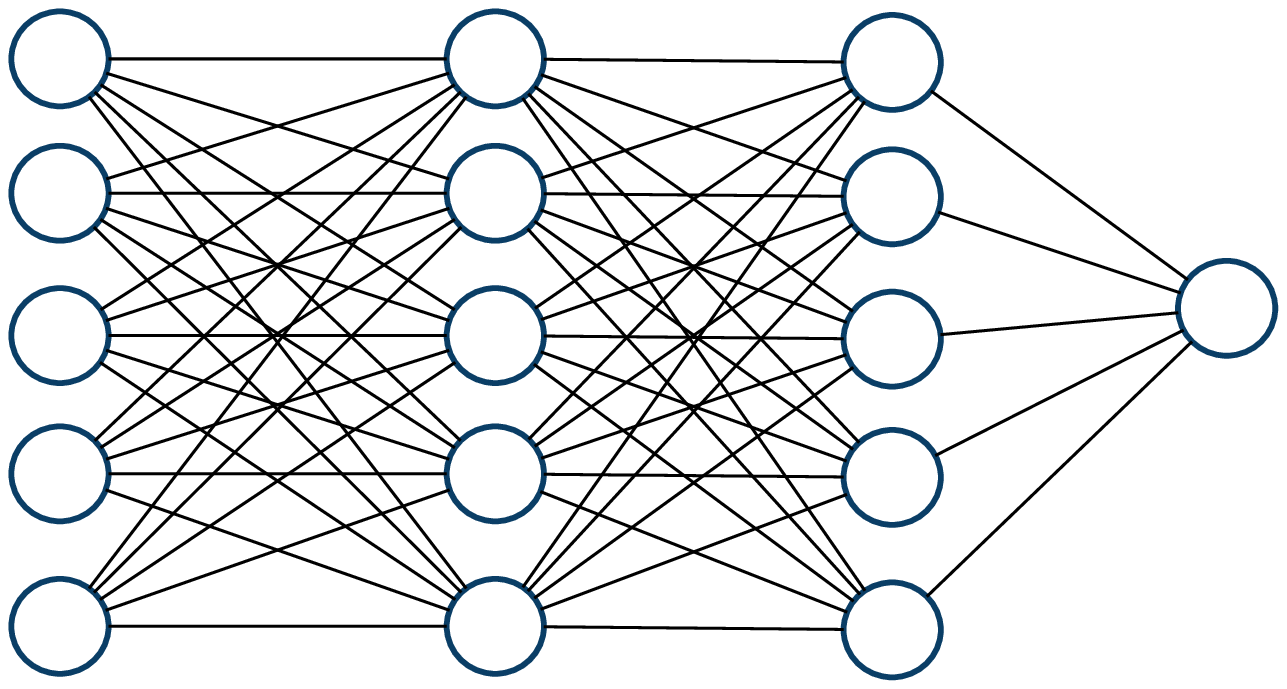}&
\includegraphics[width = 4 cm]{NN}&
\includegraphics[width = 4 cm]{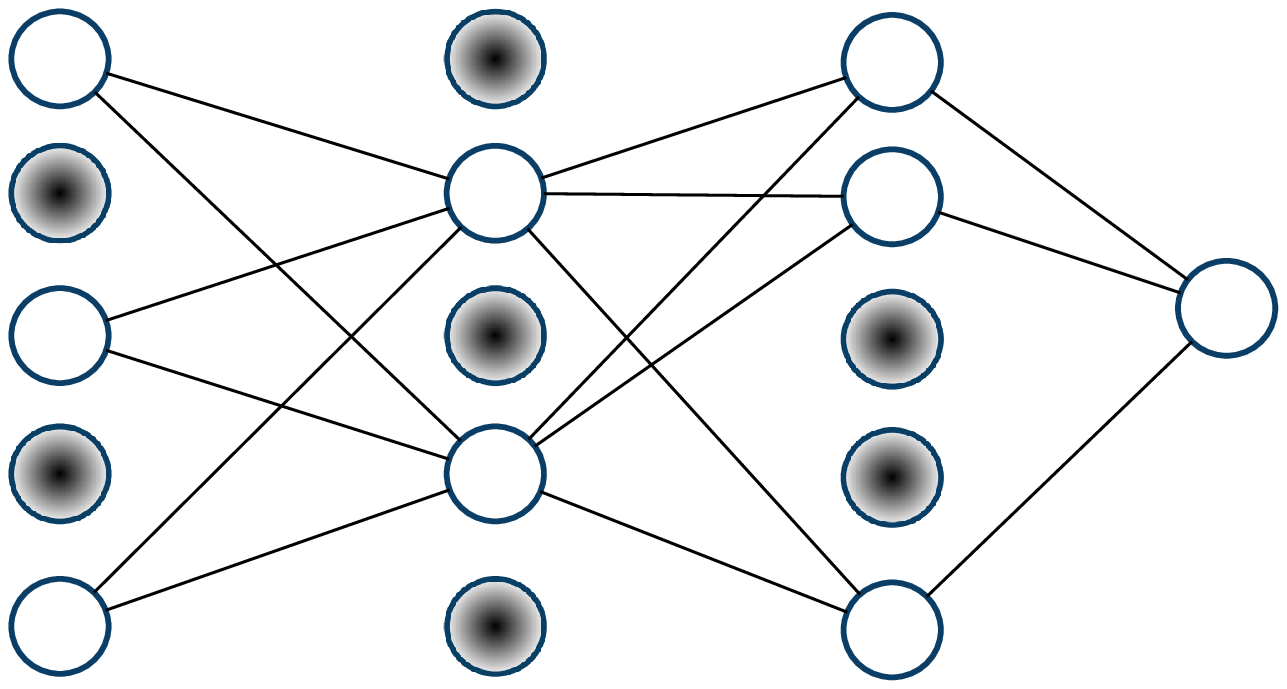}&
\includegraphics[width = 4 cm]{NN}\\
\raisebox{0 pt}[0 pt][0 pt]{\rotatebox{90}{\footnotesize Drop-Connect}}&
\includegraphics[width = 4 cm]{NN}&
\includegraphics[width = 4 cm]{NN}&
\includegraphics[width = 4 cm]{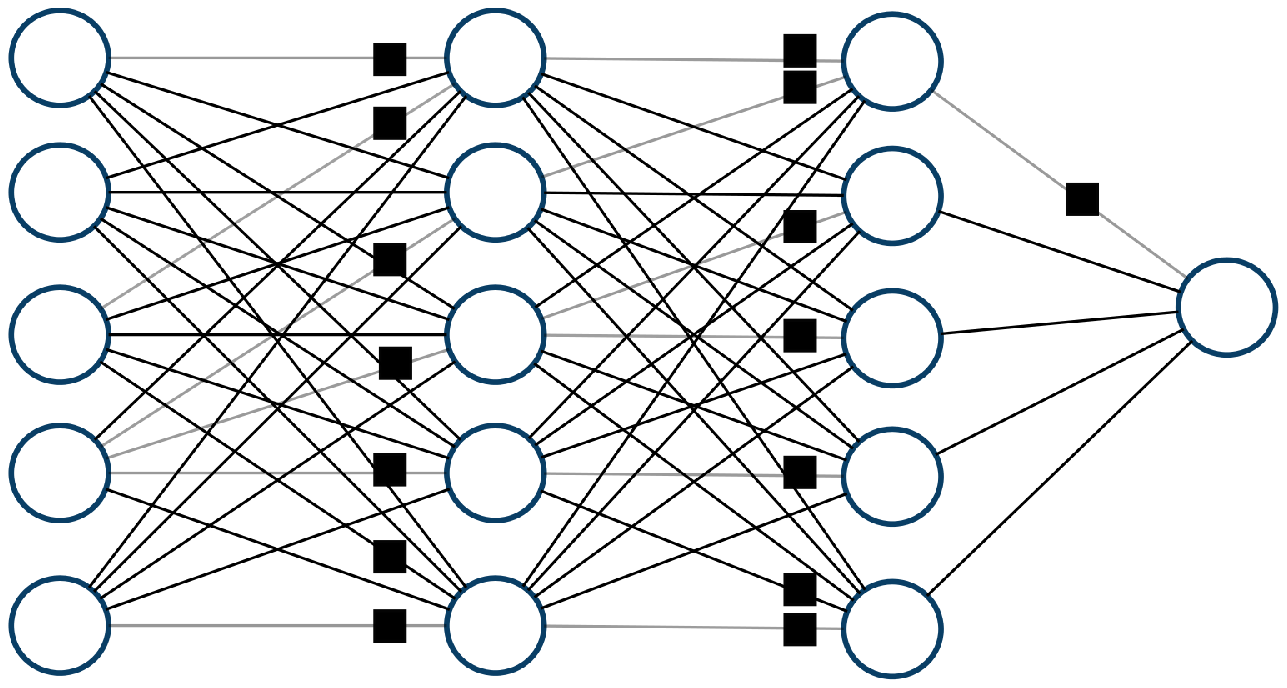}&
\includegraphics[width = 4 cm]{NN}\\
  \raisebox{8 pt}[0 pt][0 pt]{\rotatebox{90}{\footnotesize HashNet}}&
\includegraphics[width = 4 cm]{NN}&
\includegraphics[width = 4 cm]{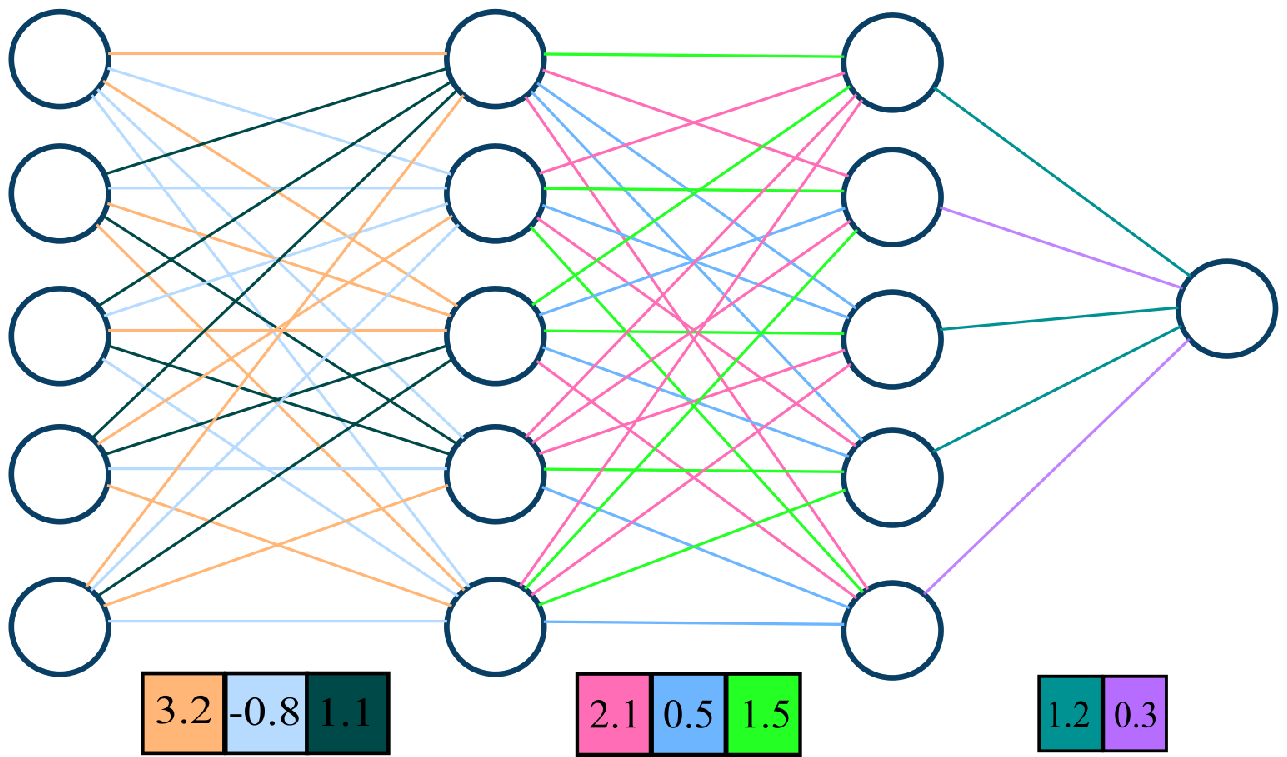}&
\includegraphics[width = 4 cm]{HashNet2}&
\includegraphics[width = 4 cm]{HashNet2}\\
  \raisebox{4 pt}[0 pt][0 pt]{\rotatebox{90}{\footnotesize StochasticNet}}&
\includegraphics[width = 4 cm]{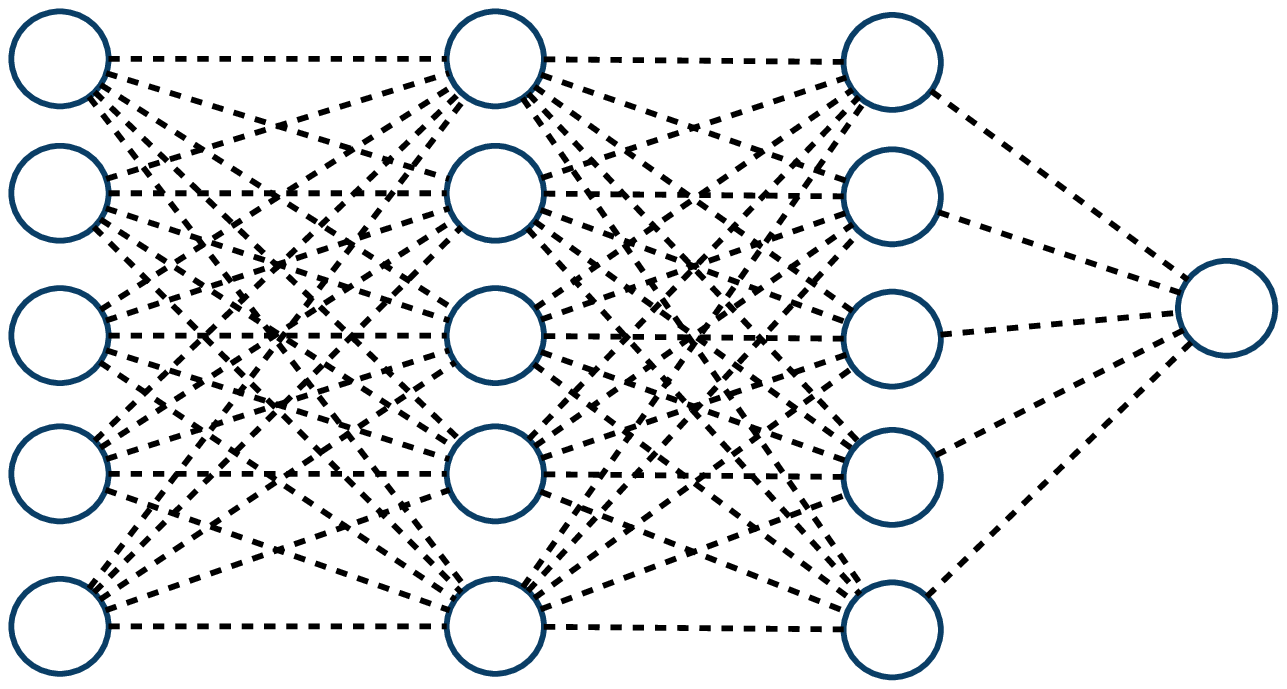}&
\includegraphics[width = 4 cm]{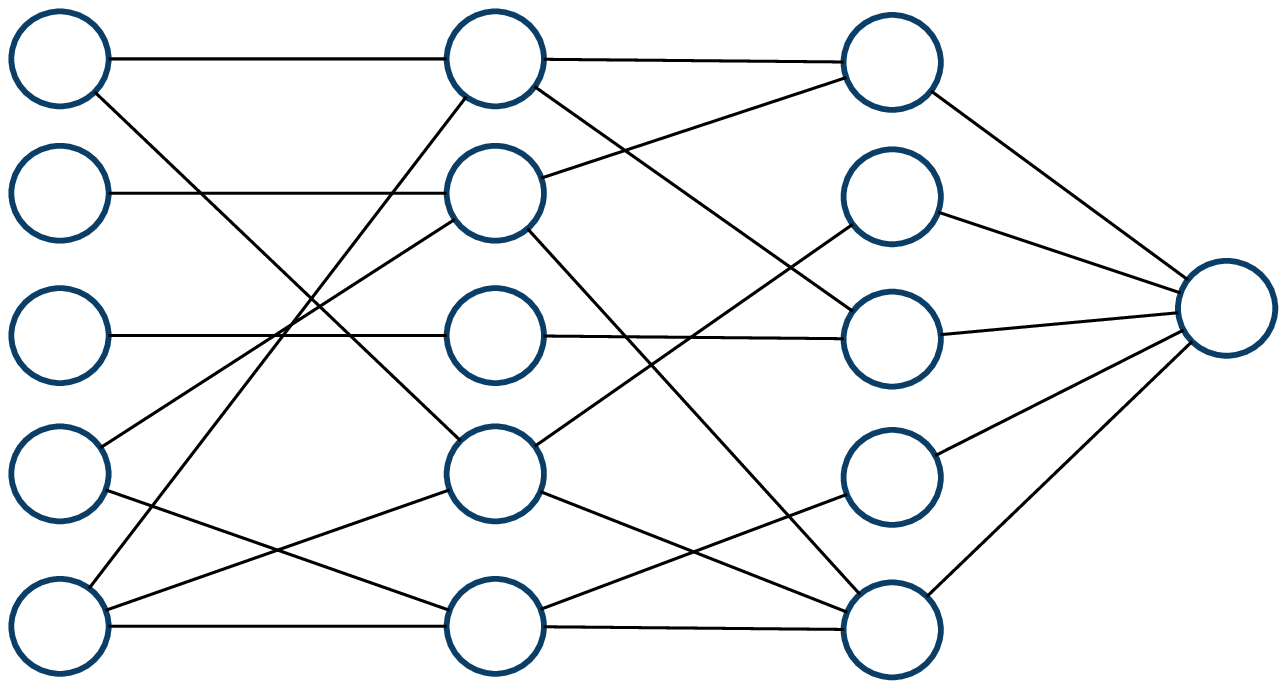}&
\includegraphics[width = 4 cm]{StochasticNetPreTrained}&
\includegraphics[width = 4 cm]{StochasticNetPreTrained}\\
\end{tabular}
\caption{ Visual comparison among different related methods to StochasticNet. Constructing a deep neural network  can be divided into 4 steps. 1) Network Model: dropout, drop-connect and Hashnet methods start with the conventional convNet where as stochastic starts with a random graph. 2) Network Formation: droput and drop-connect approaches utilize the formed network from previous step while Hashent groups edges to have same weight (shown in same color) and stochastic net samples the constructed random graph. 3)Training: dropout and drop-connect methods drop actiavation function results or the weight connectivities randomly to regularize the training network while the Hashnet and StochasticNet obtain the same configuration as previous step (i.e., Network Formation). 4) Testing: in this step dropout, drop-connect and Hashnet approximate all network parameters and get the output based on the complete network structure while StochasticNet outputs by the sparse trained network.  }
\label{fig:diffIllustration}
\end{center}
\vspace{- 0.25cm}
\end{figure*}
\section{Experimental Results}
To investigate the efficacy of efficient feature learning via StochasticNets, we form deep convolutional StochasticNets and train the constructed StochasticNets using the CIFAR-10~\cite{CIFAR10} image dataset for generating generic features.  Based on the trained StochasticNets, features are then extracted for the SVHN~\cite{SVHN} and STL-10~\cite{STL10} image datasets and image classification performance using these extracted deep features within a neural network classifier framework are then evaluated in a number of different ways.  It is important to note that the main goal is to investigate the efficacy of feature learning via StochasticNets and the influence of stochastic connectivity parameters on feature representation performance, and not to obtain maximum absolute classification performance; therefore, the performance of StochasticNets can be further optimized through additional techniques such as data augmentation and network regularization methods.

 %\vspace{- 0.5 cm}
\subsubsection{Datasets}
The CIFAR-10 image dataset~\cite{CIFAR10} consists of 50,000 training images categorized into 10 different classes (5,000 images per class) of natural scenes. Each image is an RGB image that is 32$\times$32 in size.  The MNIST image dataset~\cite{MNIST} consists of 60,000 training images and 10,000 test images of handwritten digits.  Each image is a binary image that is 28$\times$28 in size, with the handwritten digits are normalized with respect to size and centered in each image.  The SVHN image dataset~\cite{SVHN} consists of 604,388 training images and 26,032 test images of digits in natural scenes. Each image is an RGB image that is 32$\times$32 in size.  Finally, the STL-10 image dataset~\cite{STL10} consists of 5,000 labeled training images and 8,000 labeled test images categorized into 10 different classes (500 training images and 800 training images per class) of natural scenes. Each image is an RGB image that is 96$\times$96 in size. The images  were resized to $32 \times 32$ to have the same network configuration for all experimented datasets for consistency purposes.  Note that  the 100,000 unlabeled images in the STL-10 image dataset were not used in this paper.

\subsubsection{StochasticNet Configuration}
\label{SNETCONFIG}
% of size $5 \times 5$
The deep convolutional StochasticNets used in this paper are realized based on the LeNet-5 deep convolutional neural network~\cite{MNIST} architecture, and consists of three convolutional layers with 32, 32, and 64  receptive fields for the first, second, and third convolutional layers, respectively, and one hidden layer of 64 neurons, with all neural connections being randomly realized based on probability distributions.  The neural connectivity formation for the deep convolutional StochasticNet realizations is achieved via acceptance-rejection sampling, and can be expressed by:
\begin{align}
e_{i\rightarrow j} \;\; \text{exists where}\;\; \Big[p({i\rightarrow j}) \geq T  \Big] = 1
\end{align}
where $e_{i\rightarrow j}$ is the neural connectivity from node $i$ to node $j$,  $[\cdot]$ is the Iverson bracket, and $T$ encodes the sparsity of neural connectivity in the StochasticNet.
 \begin{figure*}
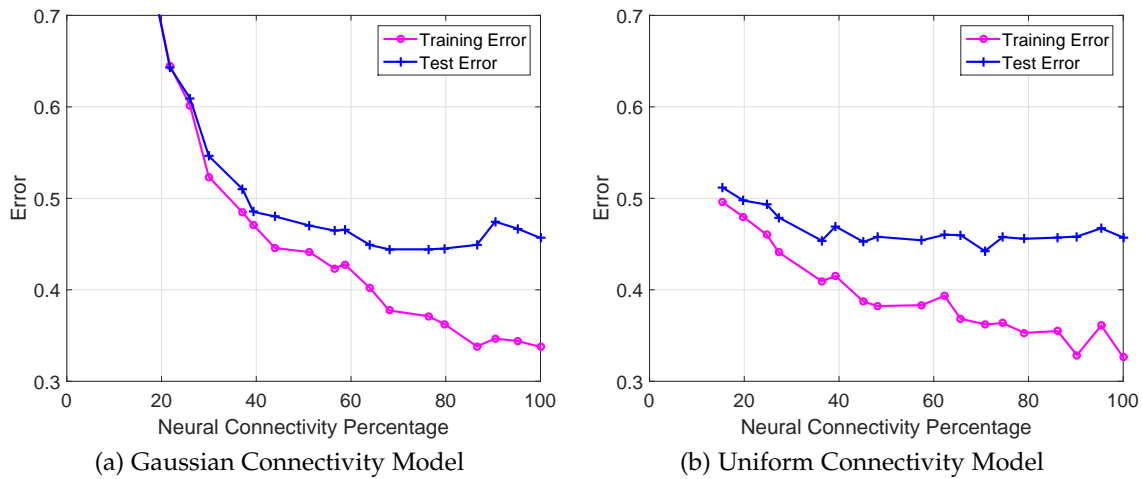

\vspace{- 1cm}
\begin{center}
\begin{tabular}{cc}
\includegraphics[scale = 0.4]{TrainingVSCon_Gauss}&
\includegraphics[scale = 0.4]{TrainingVSCon_Uniform}\\
(a) Gaussian Connectivity Model & (b) Uniform Connectivity Model
\end{tabular}
\caption{Training and test error versus the number of neural connections for the STL-10 dataset. Both Gaussian and uniform neural connectivity models were evaluated. Note that neural connectivity percentage of 100 is equivalent to ConvNet, since all connections are made. As seen a StochasticNet with fewer neural connectivity compared to the ConvNet provides better than or similar accuracy to the ConvNet.  }
\label{fig:TrainingVSCon}
\end{center}
\vspace{- 0.25cm}
\end{figure*}
While it is possible to take advantage of any arbitrary distribution to construct deep convolutional StochasticNet realizations, for the purpose of this paper two different spatial neural connectivity models were explored for the convolutional layers: \mbox{i) uniform} connectivity model:
\[
p({i\rightarrow j})= \begin{cases}
U(0,1) & j \in R_i\\
0 & \text{otherwise.}
\end{cases}
\]
 and ii) a Gaussian connectivity model:
 \[
p({i\rightarrow j})= \begin{cases}
\mathcal{N}(i,\sigma) & j \in R_i\\
0 & \text{otherwise.}
\end{cases}
\]
where the mean is at the center of the receptive field (i.e., $i$) and the standard deviation $\sigma$ is set to be a third of the receptive field size. In this study, $R_i$ is defined as a $5 \times 5$ spatial region around node $i$, which means that neural connectivity of 100 the resulting receptive field is equivalent to a dense $5 \times 5$ receptive field used for ConvNets.  Finally, for comparison purposes, the conventional ConvNet used as a baseline is configured with the same network architecture using $5 \times 5$ receptive fields.

\begin{figure*}
\vspace{-1 cm}
\begin{center}
\begin{tabular}{cc}
\includegraphics[scale = 0.4]{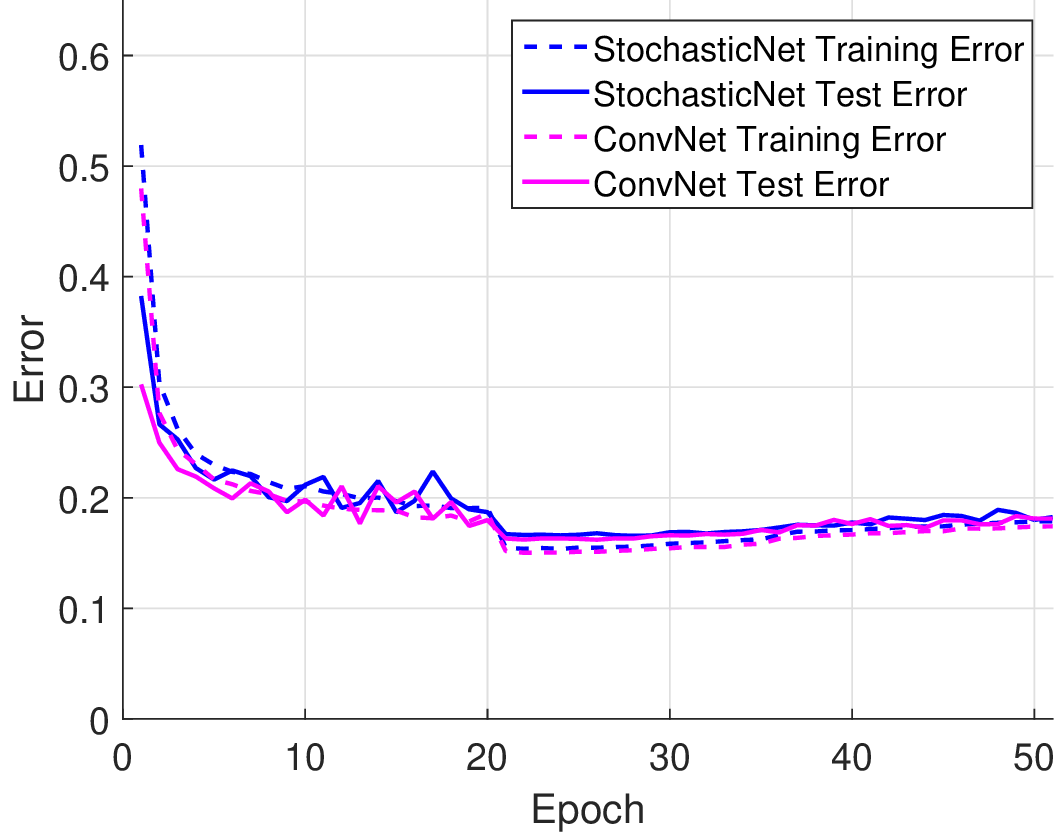}&
\includegraphics[scale = 0.4]{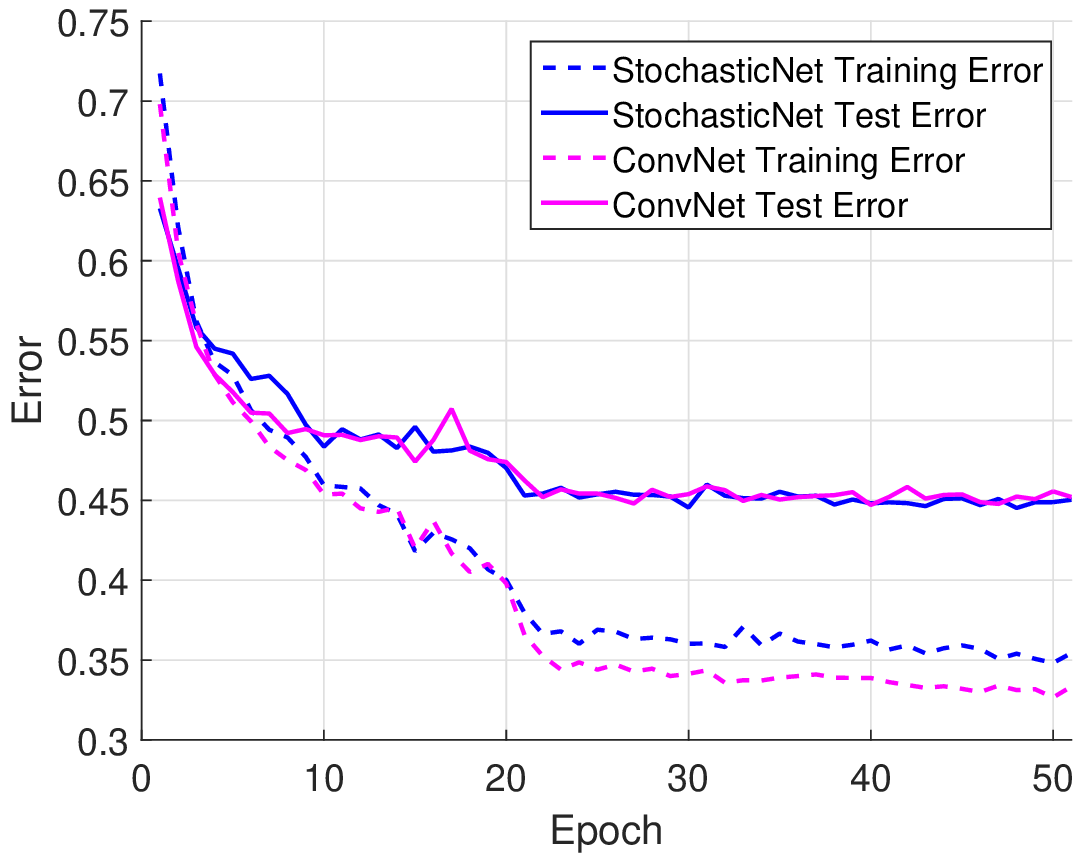}\\
(a) SVHN & (b) STL-10\\
\end{tabular}
\caption{ Comparison of classification performance between deep features extracted with a standard ConvNet and that extracted with a StochasticNet containing 75\% neural connectivity as the ConvNet. The StochasticNet is realized based on a uniform connectivity model. The StochasticNet results in a 0.5\% improvement in relative test error for the STL-10 dataset, as well as provides a smaller gap between the training error and test error. The StochasticNet using a uniform connectivity model achieved the same performance as the ConvNet for the SVHN dataset.   }
\label{fig:StochasticNetVSConvNetA}
\end{center}
 %\vspace{- 0.5cm}
\end{figure*}

 %\vspace{- 0.5 cm}
\subsection{Number of Neural Connections}
\label{numconnections}
An experiment was conducted to illustrate the impact of the number of neural connections on the feature representation capabilities of StochasticNets. Figure~\ref{fig:TrainingVSCon} demonstrates the training and test error versus the number of neural connections in the network for the STL-10 dataset.  The neural connection probability is varied to achieve the desired number of neural connections for testing its effect on feature representation capabilities.

Figure~\ref{fig:TrainingVSCon} demonstrates the training and testing error vs. the neural connectivity percentage relative to the baseline ConvNet, for two different spatial neural connectivity models: i) uniform connectivity model, and ii) Gaussian connectivity model.   It can be observed that classification using the features from the StochasticNet is able to achieve the better or similar test error as using the features from the ConvNet when the number of neural connections in the StochasticNet is fewer than the ConvNet.  In particular, classification using the features from the StochasticNet is able to achieve the same test error as using the features from the ConvNet when the number of neural connections in the StochasticNet is half that of the ConvNet.  It can be also observed that, although increasing the number of neural connections resulted in lower training error, it does not exhibit reductions in test error, and as such it can be observed that the proposed StochasticNets can improve the handling of over-fitting associated with deep neural networks while decreasing the number of neural connections, which in effect greatly reduces the number of computations and thus resulting in more efficient feature learning and feature extraction.  Finally, it is also observed that there is a noticeable difference in the training and test errors when using the Gaussian connectivity model when compared to the uniform connectivity model, which indicates that the choice of neural connectivity probability distributions can have a noticeable impact on feature representation capabilities.

\begin{figure*}[!htp]
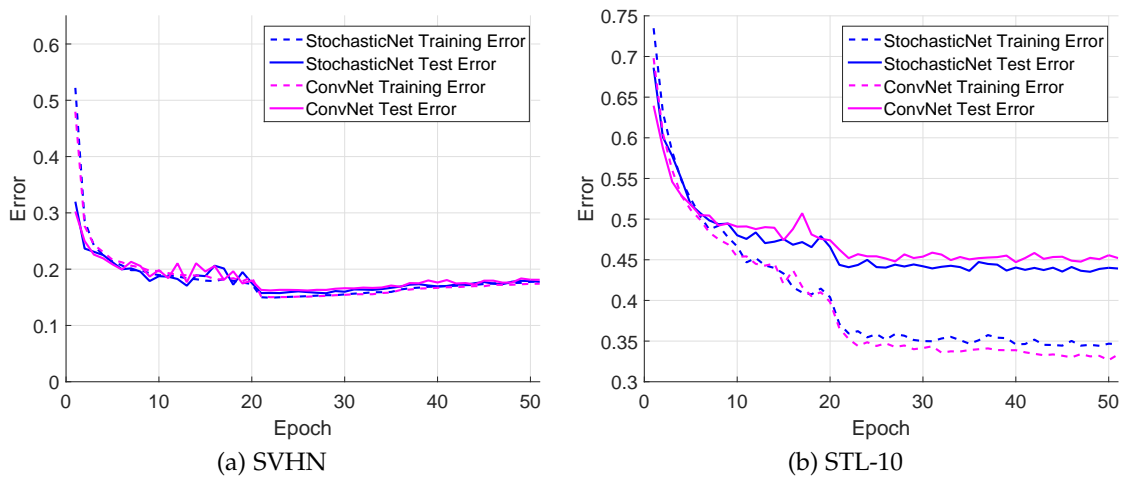

 %\vspace{- 0.5cm}
\begin{center}
\begin{tabular}{cc}
\includegraphics[scale = 0.4]{SVHNbyCifar_Gauss}&
\includegraphics[scale = 0.4]{STL10byCifar_Gaussian}\\
(a) SVHN & (b) STL-10\\
\end{tabular}
\caption{ Comparison of classification performance between deep features extracted with a standard ConvNet and that extracted with a StochasticNet containing 75\% neural connectivity as the ConvNet. The StochasticNet is realized based on a Gaussian connectivity model. The StochasticNet results in a 4.5\% improvement in relative test error for the STL-10 dataset, as well as provides a smaller gap between the training error and test error. A 1\% relative improvement is also observed for the SVHN dataset.   }
\label{fig:StochasticNetVSConvNetB}
\end{center}
  %\vspace{- 0.5cm}
\end{figure*}

 %\vspace{- 0.5 cm}
\subsection{Comparisons with ConvNet for Feature Learning}

Motivated by the results shown in Figure~\ref{fig:TrainingVSCon}, a comprehensive experiment were done to investigate the efficacy of feature learning via StochasticNets on CIFAR-10 and utilize them to classify the SVHN and STL-10 image datasets. Deep convolutional StochasticNet realizations were formed with 75\% neural connectivity using the Gaussian connectivity model as well as the uniform connectivity model when compared to a conventional ConvNet.  The performance of the StochasticNets and the ConvNets was evaluated based on 25 trials and the reported results are based on the best of the 25 trials in terms of training error. Figure~\ref{fig:StochasticNetVSConvNetA} and Figure~\ref{fig:StochasticNetVSConvNetB} shows the training and test error results of classification using learned deep features  from CIFAR-10 using the StochasticNets  and ConvNets on the SVHN and STL-10 datasets via the uniform connectivity model and the Gaussian connectivity model, respectively.  It can be observed that, in the case where the uniform connectivity model is used, the test error for classification using features learned using StochasticNets, with just 75\% of neural connections as ConvNets, is approximately the same as ConvNets for both the SVHN and STL-10 datasets (with $\sim$0.5\% test error reduction for STL-10).  It can also be observed that, in the case where the Gaussian connectivity model is used, the test error for classification using features learned using StochasticNets, with just 75\% of neural connections as ConvNets, is approximately the same ($\sim$1\% relative test error reduction) as ConvNets for the SVHN dataset.  More interestingly, it can also be observed that the test error for classification using features learned using StochasticNets, with just 75\% of neural connections as ConvNets, is reduced by $\sim$4.5\% compared to ConvNets for the STL-10 dataset. Furthermore, the gap between the training and test errors of classification using features extracted using the StochasticNets is less than that of the ConvNets, which would indicate reduced overfitting in the StochasticNets.

These results illustrate the efficacy of feature learning via StochasticNets in providing efficient feature learning and extraction while preserving feature representation capabilities, which is particularly important for real-world applications where efficient feature extraction performance is necessary.

 \begin{figure*}[!ht]
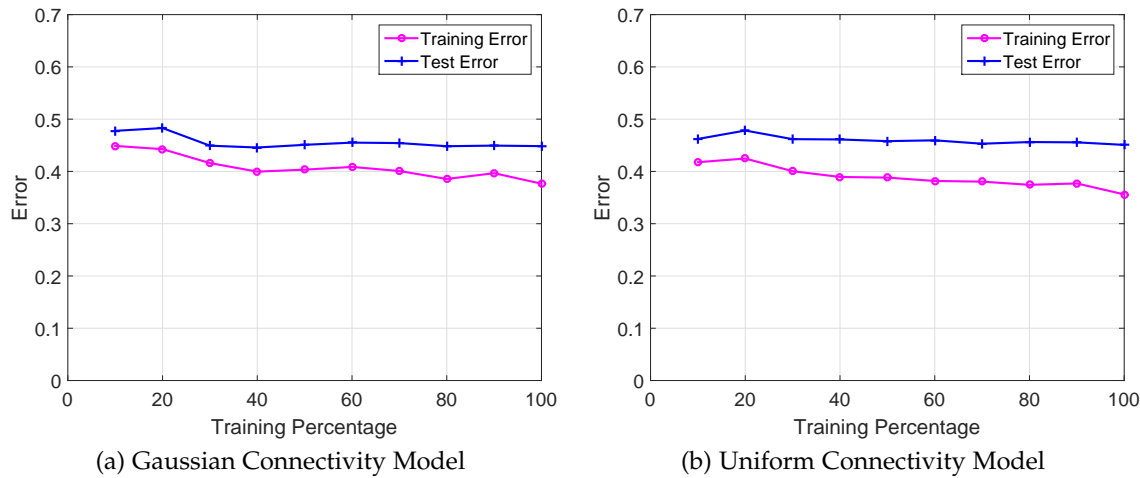

  %\vspace{- 0.5cm}
\begin{center}
\begin{tabular}{cc}
\includegraphics[scale = 0.4]{AccVSnumTrain_Gauss_STL10}&
\includegraphics[scale = 0.4]{AccVSnumTrain_uniform_STL10}\\
(a) Gaussian Connectivity Model & (b) Uniform  Connectivity Model
\end{tabular}
\caption{Training and test error for classifying the STL-10 dataset versus the training set size used to train the network from CIFAR-10 dataset. Deep convolutional StochasticNet realizations were formed with 75\% neural connectivity via the Gaussian connectivity model as well as the uniform connectivity model.  As seen the features extracted using the StochasticNets can provide comparable classification performance even when only 30\% of the training data is used in the case of the Gaussian connectivity model, and there was only a 3\% increase in test error when only 10\% of the training data is used.  More interest is the case of the uniform connectivity model, where the features extracted using the StochasticNets can provide comparable classification performance with no increase in test error even when only 10\% of the training data. }
%  %\vspace{- 1cm}
\label{fig:TrainingVSSetSize}
\end{center}
\end{figure*}
%%\vspace{- 0.5 cm}
\subsection{Training Set Size}
\label{numconnections}

An experiment was conducted to illustrate the impact of the size of the training set on the feature representation capabilities of StochasticNets.  To perform this experiment, deep convolutional StochasticNet realizations were formed with 75\% neural connectivity using the Gaussian connectivity model as well as the uniform connectivity model, and different percentages of the CIFAR-10 dataset was used for feature learning.  The trained StochasticNet realizations where then used to perform classification on the STL-10 dataset to evaluate training and test error performance analysis.

Figure~\ref{fig:TrainingVSSetSize} demonstrates the training and testing error vs. the training set size, for the two tested connectivity models.   It can be observed that the features extracted using the StochasticNets can provide comparable classification performance even when only 30\% of the training data is used in the case of the Gaussian connectivity model.  Furthermore, it was observed that there was only a 3\% drop in test error when only 10\% of the training data is used.  More interest is the case of the uniform connectivity model, where the features extracted using the StochasticNets can provide comparable classification performance with no increase in test error even when only 10\% of the training data, which illustrates the efficacy of feature learning via StochasticNets in situations where the training size is small.

\subsection{Relative Feature Extraction Speed vs. Number of Neural Connections}

Previous sections showed that StochasticNets can achieve good feature learning performance relative to conventional ConvNets, while having significantly fewer neural connections. We now investigate the feature extraction speed of StochasticNets, relative to the feature extraction speed of ConvNets, with respect to the number of neural connections formed in the constructed StochasticNets. To this end, the convolutions in the StochasticNets are implemented as a sparse matrix dot product, while the convolutions in the ConvNets are implemented as a matrix dot product.  For fair comparison, both implementations do not make use of any hardware-specific optimizations such as Streaming SIMD Extensions (SSE) because many industrial embedded architectures used in applications such as embedded video surveillance systems do not support hardware optimization such as SSE.

\begin{figure}
\vspace{- 0.5cm}
\begin{center}
\begin{tabular}{cc}
\includegraphics[scale = 0.25]{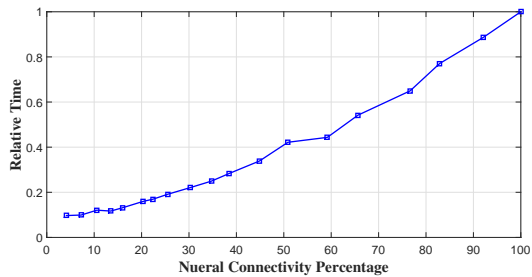}
\end{tabular}
\caption{Relative feature extraction time versus the number of neural connections. Note that neural connectivity percentage of 100 is equivalent to ConvNet, since all connections are made.}
\label{fig:SpeedVSCon}
\end{center}
\vspace{- 0.5cm}
\end{figure}

As with Section~\ref{numconnections}, the neural connection probability is varied in both the convolutional layers and the hidden layer to achieve the desired number of neural connections for testing its effect on the feature extraction speed of the formed StochasticNets.  Figure~\ref{fig:SpeedVSCon} demonstrates the relative feature extraction time vs. the neural connectivity percentage, where relative time is defined as the time required during the feature extraction process relative to that of the ConvNet.  It can be observed that the relative feature extraction time decreases as the number of neural connections decrease, which illustrates the potential for StochasticNets to enable more efficient feature extraction.

Interestingly, it can be observed that speed improvements are seen immediately, even at 90\% connectivity, which may appear quite surprising given that sparse representation of matrices often suffer from high computational overhead when representing dense matrices.  However, in this case, the number of connections in the randomly realized receptive field is very small relative to the typical input image size. As a result, the overhead associated with using sparse representations is largely negligible when compared to the speed improvements from the reduced computations gained by eliminating even one connection in the receptive field.  Therefore, these results show that StochasticNets can have significant merit for enabling the feature representation power of deep neural networks to be leveraged for a large number of industrial embedded applications.

% %\vspace{- 0.5cm}
\section{Conclusions}
%--------------------------------------------------------------------------------------------------------------------------------------------------

In this paper, we proposed the learning of efficient feature representations via StochasticNets, where sparsely-connected deep neural networks are constructed by way of stochastic connectivity between neurons.  Such an approach facilitates for more efficient neural utilization, resulting in reduced computational complexity for feature learning and extraction while preserving feature representation capabilities. The effectiveness of feature learning via StochasticNet was investigated by training StochasticNets using the CIFAR-10 dataset and using the learned features for classification of images in the SVHN and STL-10 image datasets. The StochasticNet features were then compared to the features extracted using a conventional convolutional neural network (ConvNet). Experimental results demonstrate that classification using features extracted via StochasticNets provided better or comparable accuracy than features extracted via ConvNets, even when the number of neural connections is significantly fewer. Furthermore, StochasticNets, with fewer neural connections than the conventional ConvNet, can reduce the over fitting issue associating with the conventional ConvNet.  As a result, deep feature learning and extraction via StochasticNets can allow for more efficient feature extraction while facilitating for better or similar accuracy performances.

\vspace{-0.5 cm}
\ifCLASSOPTIONcompsoc
  \section*{Acknowledgment}
\else
  \section*{Acknowledgment}
\fi

This work was supported by the Natural Sciences and Engineering Research Council of Canada, Canada Research Chairs Program, and the Ontario Ministry of Research and Innovation. The authors also thank Nvidia for the GPU hardware used in this study through the Nvidia Hardware Grant Program.

\vspace{-0.25 cm}

\ifCLASSOPTIONcaptionsoff
  \newpage
\fi



%\bibliographystyle{IEEEtran}
 %\vspace{-0.3 cm}
 \bibliographystyle{IEEEtran}
\bibliography{refs}
%\begin{thebibliography}{1}
%H.~Kopka and P.~W. Daly, \emph{A Guide to {\LaTeX}}, 3rd~ed.\hskip 1em plus
%\end{thebibliography}


%\begin{IEEEbiography}{Mohammad Javad Shafiee}
%\end{IEEEbiography}
%\begin{IEEEbiographynophoto}{Alexander Wong}
%% insert where needed to balance the two columns on the last page with
%%\newpage
%
%\begin{IEEEbiographynophoto}{Paul Fieguth}


%\vfill


\end{document}